\pdfoutput=1
\documentclass[11pt,a4paper]{article}
\usepackage{authblk}
\usepackage[hyperref]{acl2020}
\usepackage{times}
\usepackage{latexsym}

\usepackage{microtype}
\usepackage{graphicx}
\usepackage{bm}
\aclfinalcopy % Uncomment this line for the final submission
\title{Open Set Authorship Attribution toward Demystifying Victorian Periodicals}

\author[1]{Sarkhan Badirli}
\author[2]{Mary Borgo Ton}
\author[3]{Abdulmecit Gungor}
\author[4]{Murat Dundar}
\affil[1]{Computer Science Department, Purdue University, Indiana, USA}
\affil[2]{Indiana University Libraries, Indiana University, Indiana, USA}
\affil[3]{Intel, Oregon, USA}
\affil[4]{Computer and Information Science Department, IUPUI, Indiana, USA}
\affil[ ]{\href{mailto:sbadirli@iu.edu}{sbadirli@purdue.edu}, \href{mailto:meborgo@indiana.edu}{meborgo@indiana.edu}, \href{mailto:abdulmecit.gungor@intel.com}{abdulmecit.gungor@intel.com},\href{mailto:mdundar@iupui.edu}{mdundar@iupui.edu}}
\date{}

\begin{document}
\maketitle
\begin{abstract}
Existing research in computational authorship attribution (AA) has primarily focused on attribution tasks with a limited number of authors in a closed-set configuration. This restricted set-up is far from being realistic in dealing with highly entangled real-world AA tasks that involve a large number of candidate authors for attribution during test time. In this paper, we study AA in historical texts using a new data set compiled from the Victorian literature. We investigate the predictive capacity of most common English words in  distinguishing writings of most prominent Victorian novelists. We challenged the closed-set classification assumption and discussed the limitations of standard machine learning techniques in dealing with the open set AA task. Our experiments suggest that a linear classifier can achieve near perfect attribution accuracy under closed set assumption yet, the need for more robust approaches becomes evident once a large candidate pool has to be considered in the open-set classification setting. %Finally, we tested our approach on pseudonymous essays from Victorian Periodicals and compared our findings with a traditional AA approach.
\end{abstract}

\section{Introduction}

Deriving its name from  Queen Victoria (1837-1901) of Great Britain, Victorian literature encompasses some of the most widely-read English writers, including Charles Dickens, the Brontë sisters, Arthur Conan Doyle and many other eminent novelists. The most prolific Victorian authors shared their freshest interpretations of literature, religion, politics, social science and political economy in monthly periodicals (W. F. Poole, 1882).  
To avoid damaging their reputations as novelists, almost ninety percent of these articles were either written anonymously or under pseudonyms. As the pseudonym gave authors greater license to express their personal, political, and artistic views, anonymous essays were often honest and outspoken, particularly on controversial issues. However, this publication strategy presents literary critics and historians with a significant challenge. %For an insightful, reasonably accurate, and rational interpretation of a Victorian article, knowing the identity of the author and his/her previous work has vital importance. 
As Houghton notes, knowing the author's identity can radically reshape interpretations of anonymously-authored articles, particularly those that include political critique. %(This concrete example was given by the editor of Wellesley:  An anonymous paper attacking the Thirty-nine Articles would mean one thing if it were written by T. H. Huxley and something quite different if the author were the Bishop of London)
Furthermore, the intended audience of an anonymous essay cannot be accurately identified without knowing the true identity of the contributor (Walter Houghton, 1965).

To address this problem, %Until 1965 and Wellesley index, there wasn't any author indexing study to tackle this problem. 
Walter Houghton %of University of Toronto 
worked in collaboration with staff members, a board of editors, librarians and scholars from all over the world to pioneer the traditional approach to authorship attribution. In 1965, they created a 5-volume journal of \textit{Wellesley Index to Victorian Periodicals}, named after Houghton's  during his time in Wellesley College. Since then additions and corrections to the  \citet{wellesley_index} Index are recorded in the \citet{curran_index} Index. Together, the Wellesley and the Curran Indices have become the primary resources for author indexing in Victorian periodicals. 

In his introduction to the \citet{wellesley_index} Index, the editor in chief Dr. Houghton describes the main sources of evidence used for author indexing. "In making the identifications, we have not relied on stylistic characteristics, it was external evidence: passages in published biographies and letters, collections of essays which are reprints of anonymous articles, marked files of the periodicals, publishers' lists and account books, and the correspondence of editors and leading contributors in British archives." Such an approach draws from the disciplinary strengths of historically-oriented literary criticism. However, this method of author identification cannot be used on a larger scale, for this approach requires a massive amount of human labor to track all correspondence related to published essays, the vast majority of which appear only in manuscript form or in edited volumes in print. Furthermore, the degree of ambiguity in these sources, combined with the incomplete nature of correspondence archives, makes a definitive interpretation difficult to achieve. 

Our study focuses on the stylistic characteristics previously overlooked in these indices. Even though author attribution using stylistic characteristics may become impractical when the potential number of contributors is on the order of thousands, we demonstrate that attribution with a high degree of accuracy is still possible based on usage and frequency of the most common words as long as one is interested in the works of a select few contributors with an adequate number of known work available for training. To support this, we focused our study on 36 of the most eminent and prolific writers of the Victorian era who have 4 or more published books that are accessible through Project \citet{gutenberg}.  However, the fact that any given work can belong to one of the few select contributors as well as any one of the thousands of potential contributors still poses a significant technical challenge for traditional machine learning, especially for Victorian periodical literature. Less than half of the articles in these periodicals were written by  well-known writers; the rest was contributed by thousands of less known individuals many of those first occupation is not literature. The technical challenges posed by this diverse pool of authors can only be addressed with an open-set classification framework. 

Unlike its traditional counterpart, computational Authorship Attribution (AA) identifies the author of a given text out of possible candidates using statistical/machine learning methods. Thanks to the seminal work of \citet{mostellar:64} in the second half of the $20$th century, AA has become one of the most prevalent application areas of modern Natural Language Processing (NLP) research. With the advances in internet and smartphone technology, the accumulation of text data has accelerated at an exponential rate in the last two decades. Abundant text data available in the form of  emails, tweets, blogs and electronic version of books \citep{gutenberg, gdelt} has opened new venues for authorship attribution and led to a surge of interest in AA among academic communities as well as corporate establishments and government agencies. 

Most of the large body of existing work in AA utilizes methods that expose subtle stylometric features. These methods include n-grams, word/sentence length and vocabulary richness. When paired with word distributions, these approaches have achieved promising results performing various AA tasks. This paper explores deviates from this trend by asking a simple yet pressing question: Can we identify authors of texts written by the world's most prominent writers based on the distributions of the words most commonly used in daily life? As a corollary, which factors play the central role in shaping predictive accuracy? Toward achieving this end, we have investigated writings of most renowned Victorian era novelists by quantifying their writing patterns in terms of their usage frequency of the most common English words. We investigated the effect of different variables on the performance of the classifier model  to explore the strengths and limitations of our approach. The experiments are run in an open-set classification setup to reflect real world characteristics of our multi-authored corpus. %This is achieved by integrating an outlier detection capability to a standard supervised classifier model. 
Unattributed articles in this corpus of Victorian texts are identified as belonging to one of the 46 known novelists or classified as an unknown author. Results are evaluated using Wellesley index as a reference standard.

\section{Related Work}

Computational AA has offered compelling analyses of well-known documents with unknown or disputed attribution. AA has contributed to conversations about the authorship of the disputed Federalist papers \cite{mostellar:64}, the Shakespearean authorship controversy \cite{shakespeare_dispute}, the author of New Testament \cite{new_testament:13} and the author of The Dark Tower \cite{dark_tower:13} to name a few examples. AA has also been proven to be very effective in forensic linguistic science. Two noteworthy examples of scholarship in this vein include the use of CUSUM (Cumulative Sum Analysis) technique \cite{morton_et_al:90} as expert evidence in court proceedings and the use of stylometric analysis by FBI agents in solving the \textit{unabomber} case \cite{unibomber}.

Beyond its narrow application within literary research, AA has also paved the way for the development of several other tasks \cite{Stamatatos:09} such as author verification \cite{koppel_schler:04}, plagiarism detection \cite{plagiarism_detection:07}, author profiling \cite{author_profiling:02}, and the detection of stylistic inconsistencies \cite{stylistic_incost:04}.

\citet{mostellar:64} pioneered the statistical approach to AA by using functional and non-contextual words to identify authors of disputed "Federalist Papers,"  written by three American congressmen. Following their success with basic lexical features, AA as a subfield of research was dominated by efforts to define  stylometric features \cite{holmes:98, rudman:98}. Over time, features used to quantify writing style, i.e., style markers, \cite{Stamatatos:09} evolved from simple word frequency \cite{burrows:92} and word/sentence length \cite{mendenhall:87}  to more sophisticated syntactic features like sentence structure and part-of-speech tags \cite{baayen_etal:96, argamon_etal:98}, to more contextual character/word n-grams \cite{deVel_etal:01, peng_etal:04}  and finally towards semantic features like synonyms \cite{McCarthy_etal:06} and semantic dependencies \cite{gamon:04}. 

As \citet{Stamatatos:09} points out, the more detailed the text analysis for extracting stylometric features is, the less accurate and noisier produced measures get. Furthermore, sophisticated and contextual features are not robust enough measures to counteract the influence of topic and genre \cite{julian_etal:17}. %Indeed, models with these features might perform well in a closed-set classification setting get an edge over seen data yet they fail to well generalize to unseen data as well as to provide insight towards results obtained.

With the widespread use of internet and social media, AA research has shifted gears in the last two decades towards texts written in everyday language. The text data from these sources are more colloquial and idiomatic and much shorter in length compared to historical texts, making stylometric features such as word and character n-grams more effictive measures for AA research. Thanks to the abundance of short texts available in the form of tweets, messages, and blog posts, the most recent AA studies of shorter texts rely on various deep learning methods  \cite{koppel:14,rhodes:15,bagnall:15}. 

%Vector representation of words \cite{word2vec} is another popular technique With the emergence of vector representation available for millions of words \cite{word2vec}, Deep Learning methods \cite{dl_aa:16} are utilized to perform AA, yet combining these vectors for given paragraph or documents remained heuristic.

%Although computational AA has a history of  more than half a century, almost all the work was done in closed-set classification setup whereby all test samples are assumed to come from one of the training classes. Open set recognition \cite{openset_recognition:12} does not put any limitation for the true class of test sample, thus renders a more realistic scenario and more challenging task. In this work, we perform AA in this open-set 

%the first part of this period was loaded with comparison of different methods as well as analysis on variables affecting model performance \cite{koppel:13}. In the second half, studies are clustered around deep learning methods to tackle AA on shorter texts \cite{koppel:14}. In particular, applying CNN \cite{rhodes:15} and (multi-headed) RNN \cite{bagnall:15} on word and character n-grams,   promising results are achieved on both short and long texts. 

    In this study we demonstrate that a simple linear classifier trained with word counts from the most frequently used words alone performs surprisingly well in a highly entangled AA problem involving 36 prominent and prolific writers of the Victorian era. The success of this study depended on both training and test sets including the same set of authors. We investigate the effect of vocabulary size and the size of text fragments on the overall attribution accuracy. Existing AA research on historical texts is limited for two reasons. First, studies are limited to a small number of attribution candidates, and second  classification is performed in a closed-set setting. In order for computational AA to propose an alternative to costly and demanding manual author indexing methods and possibly challenge previous identifications of authorship, effective open-set classification \cite{openset_recognition:12} strategies are required. Our study is the first to tackle AA in an open-set classification setting. Although its attribution accuracy is far from being ideal when compared to identifications from the Wellesley Index, our approach demonstrates that a linear classifier when run in an open-set classification setting can produce interesting insights about pseudonymous and anonymous contributions to Victorian periodicals that would not be readily attainable by manual indexing techniques.

%In this work, we preferred a traditional machine learning model that delivers more interpretable results as the main goal here is to investigate and provide insight to Victorian Era Literature through the computational AA.
%Existing AA research  on historical texts is limited on two fronts. First, a very small number of attribution candidates is usually considered. Second, classification is performed in a closed-set setting. Another factor that hinders the progress in AA research on historical texts is the lack of a benchmark dataset  to fairly compare different methods. In this study, we compiled and made public an extensive literary research dataset from Project \citet{gutenberg} covering works of the most eminent and prolific 46 Victorian Era novelists for AA analysis. We also collected a small subset of articles from Victorian Periodicals to be used in validating AA techniques in a real-world open-set classification problem.

\section{Datasets and Methods}

\subsection{Datasets}
Herein, we discuss the datasets used in this study. The main dataset consists of 595 novels (books) from 36  Victorian Era novelists. We also added 23 novels to the corpus from another 10 authors who have less than 4 books available in Project \citet{gutenberg}. Novels from these 10 authors were not used during training phase. We used these novels as samples from unknown authors, noted as out-of-distribution (OOD) samples, in the open set classification experiments.  All books are downloaded from Project \citet{gutenberg}. These texts represent a wide range of genres, spanning from Gothic novels, historical fiction, satirical works, social-problem novels, detective stories, and realist novels. The distribution of genres among authors is quite erratic as some writers like Charles Dickens explored all of these genres in their novels whereas some others such as A. C. Doyle fall mainly within the purview of a single genre. Table  \ref{tab:author_distribution}  shows all 46 authors and the number of novels written by them in the dataset. This dataset is used in both closed and open-set classification experiments.

We  also curated a small collection of essays from Victorian periodicals that were republished in \textit{Bentley's Miscellaneous} (1937-1938, Volume 1 and 2 based on availability from \citet{gutenberg} Project) to test our model in a realistic open-set classification setup. We picked essays that were either written anonymously or under pseudonym. There are a total of 27 essays written under pseudonyms and 3 articles published anonymously. The length of essays ranges between 50-600 sentences. For validation we used identifications from Wellesley Index as reference standard.  
In preparing this dataset we ensured that all text not belonging to the author such as footnotes, preface, author biography, commentary etc. were all removed. Upon completion of the review process all datasets and \textit{Python Notebooks} prepared for this project  will be released through \textit{github}. 

\iffalse
\begin{figure*}
    \centering
    \includegraphics[width=16cm,height=8cm]{figures/Author_distribution.png}
    \caption{Caption}
    \label{fig:my_label}
\end{figure*}
\fi

\begin{table*}
\centering
\begin{tabular}{l|l}
 \hline
 \textbf{\# novels} & \textbf{Authors} \\
 \hline \hline
 $41 - 53$ & Margaret Oliphant, H. Rider Haggard, Anthony Trollope\\ \hline
 $31 - 40$ & Charlotte M. Yonge, George MacDonald \\ \hline
 $21 - 30$ & Bulwer Lytton, Frederick Marryat, Mary E. Braddon, Mary A. Ward, George Gissing \\ \hline
 $11-20$ & Arthur C. Doyle, Geroge Meredith, Harrison Ainsworth, Charles Dickens, Marie Corelli, \\
 & Mrs. Henry Wood, Benjamin Disraeli, Wilkie Collins, Ouida, Thomas Hardy, \\ 
 & Robert L. Stevenson, Walter Besant, R. D. Blackmore, William Thackeray, Charles Reade \\ \hline
 $4-10$ & Bram Stoker, George Eliot, Mary Cholmondeley, Elizabeth Gaskell, Frances Trollope, \\
 & George Reynolds, Charlotte Brontë, Dinah M. Craik, Catherine Gore, Ford Madox, \\ 
 & Rhoda Broughton \\ \hline
 $1 - 3$ & Eliza L. Lynton, Lewis Caroll, Samuel Butler, Sarah Grand, Thomas Hughes, \\
 & Anna Thackeray, Mona Caird, Anne Brontë, Walter Pater, Emily Brontë \\
 \hline
\end{tabular}
\caption{ There are total of 46 authors in the corpus. Of these, the most prolific 36 authors were used during the training phase of this study. Novels from the remaining 10 authors were used as OOD samples. First column represent ranges for the number of novels. Authors whose number of novels in the corpus that fall in the range are in the next column.  }
\label{tab:author_distribution}
\end{table*}

\subsection{Methods}

Our main motivation in this study is to demonstrate that simple classification models using just word counts from the most commonly used English words can be effective even in a highly entangled AA problem as long as both training and test sets come from the same set of authors. We emphasize that a closed-set classification setting is far from being realistic and is unlikely to offer much benefit in real-world AA problems that often emerge in open-set classification settings. We draw attention to the need for more sophisticated techniques for AA by extending the highly versatile linear support vector machine (SVM) beyond its use in closed-set text classification \cite{ ollson:08, dark_tower:13, argamon:05, bozkurt:07, kim_etal:10, mecit_thesis:18} to open-set classification. We show its strengths and limitations in the context of an interesting AA problem that requires identifying authorship information of anonymous and pseudonymous essays in Victorian periodicals. Although our discussion of methodology is limited to SVM, the conclusions we draw are not necessarily method specific and apply to other popular closed-set classification techniques (random forest, naive Bayes, multinomial etc.) as well. 

Linear SVM optimizes a hyperplane that maximizes the margin between the positive and negative classes while minimizing the training loss incurred by samples that fall on the wrong side of the margin. We extend SVM to open-set classification as follows. For each training class (author) we train a linear SVM with a quadratic regularizer and hinge loss in a one vs. all setting by considering all text from the given author as positive and text from all other authors as negative. During the test phase, test documents are processed by these classifiers, and final prediction is rendered based on the following possibilities. If the test sample is not classified as positive by any of the classifiers, it is labeled as belonging to an unknown author. If the test sample is classified as positive by one or more classifiers, the document is attributed to the author whose corresponding classifier generates the highest confidence score. 

The regularization parameter for each one vs. all classifier is tuned on the hold-out validation set to optimize $F_{1}$ score. $SGD\_classifier$ from python library $sklearn.linear\_model$ is used for SVM implementation. Class weight is set to $balanced$ to deal with class imbalance and for reproducibility purposes random state is fixed at $42$ (randomly chosen).

\section{Experiments}

In the first part of the experiment we demonstrate that usage frequency of most common words can offer important stylistic cues about the writings of renowned novelists. Towards this end, we utilized bag of words (BOW) representation to vectorize our unstructured text data.  We limit the vocabulary with the most common thousand words to avoid introducing topic or genre specific bias into AA problem. One third of these words are \textit{stop words} that would normally be removed in a standard document classification task. All special names/entities and punctuation are removed from the vocabulary. All words are converted to lower case.

Each book is divided into chunks of same number of sentences. Each of these text chunks is considered a document. Authors with less than 4 books/novels (10 of the 46) are considered as unknown authors in open-set classification experiments. These ten authors are randomly split between validation and test sets. Documents from the remaining 36 authors are split into three as train, validation, and test set using the  ratio of 64/16/20. At the book level all three sets are mutually disjoint. That is, all documents from the same book are used in one of the three sets. 

\subsection{Varying document length in both training and test sets}

In this experiment we investigate the size of each document on the authorship attribution performance. We considered five different document lengths, noted as $|D|$: 10, 25, 50, 100 sentences or the entire book. Python's NLTK library is used for sentence and word tokenization. The length of novels in terms of the number of sentences ranges from $390$ to $25,612$ and has a median of $6,801$. %an average length of $7,130$ and a standard deviation of $4,041$. 
A document with 50 sentences has a median of $1,142$ tokens.
%on the average $1,194$ tokens with a standard deviation of $340$.
Vocabulary size is another variable considered pair wise with document length. For vocabulary, noted as $|V|$, we considered the most frequent $100$, $500$, $1000$, $5000$, and $10000$ words. 

Table \ref{tab:doclen_exp} presents  mean $F_1$, noted as $\bar{F}_1$, scores for each pair of document length and vocabulary size. Not surprisingly the AA prediction performance improves as the document length increases.  Rate of improvement is more significant for shorter document lengths. Using entire novel as a document slightly hurts the performance as the number of training samples per class dramatically decreases. The same situation is also observed with the size of the feature vector. The rate of improvement is more remarkable for smaller vocabulary sizes. These results suggest that a vocabulary size of around 1000 and a document length of around 50 sentences can be sufficient to distinguish among eminent novelists with near perfect accuracy if three or more of  their books are available for training. 

\subsection{Varying document length in test set while keeping it fixed in training} To simulate a scenario, where documents during test time may emerge with arbitrary lengths, we fixed document length to 50 sentences in the training phase and varied the number of sentences in test samples. The very first problem in this setting is the scaling problem between test and training documents due to different document lengths. To address this problem we normalized word counts in each document by dividing each count by the maximum count in that document. This scaling operation is followed by maximum absolute value scaling (\textit{MaxAbsScaler}) on columns.  MaxAbsScaler is preferred over min-max scaling to preserve sparsity as the former does not shift/center the data. Results of this experiment is reported in Table \ref{tab:test_doclen_exp50}. %SVM is not scale invariant thus scaling has a huge effect on both predictive performance and computational efficiency. With scaling model gained five to ten notch increase over mean F1 and run four times faster.

Results in Table \ref{tab:test_doclen_exp50} suggest that using same document length for training and test sets is not necessary. Indeed, fixing document length in the training set to a number large enough to ensure documents are sufficiently informative for the classification task at hand while small enough to provide adequate number of training samples for each class yields results comparable to those achieved by varying sizes of training document length. It is interesting to note that the larger number of training samples available in this case led to improvements in the book-level attribution accuracy as all 131 test books are correctly attributed to their true authors for vocabulary size 500 or larger.

 %Indication is true for both small and large documents.  For example, with $1,000$ features mean F1 score for test samples with 10 sentences improved 2 notches from $0.48$ to $0.50$ and classification of entire books jumped 4 notches to a perfect score. All 131 test books are correctly attributed to their true authors.

\begin{table}
\centering
%\resizebox{\linewidth}{!}{%
\begin{tabular}{l|ccccc}
\textbf{$|D|$ / $|V|$} & \textbf{100} & \textbf{500} & \textbf{1K} & \textbf{5K} & \textbf{10K} \\ \hline
\textbf{Whole Book} &  0.94    &  0.95  &  0.96 & 0.98 & 0.97  \\
\textbf{100 Sent} &   0.78   &  0.96  & 0.98  & 0.99 &  1.00 \\
\textbf{50 Sent} &   0.58   &  0.88  &   0.93 & 0.98 & 0.98 \\
\textbf{25 Sent} &  0.37  &  0.70   &  0.79 & 0.90 & 0.92 \\
\textbf{10 Sent} &  0.15  &  0.39   &  0.48 & 0.64 & 0.68 
\end{tabular}
%}
\caption{\label{tab:doclen_exp} $\bar{F}_1$ scores from closed set experiment with 36 authors. Both training and test documents have the same length in terms of the number of sentences. Numbers of sentences considered are $10$, $25$, $50$, $100$ and entire book. Columns represent different vocabulary sizes. $100$, $500$, $1000$, $5000$ and $10000$ most frequent words are considered.}
\end{table}

\begin{table}
\centering
\begin{tabular}{l|ccccc}
\textbf{$|D|$ / $|V|$} & \textbf{100} & \textbf{500} & \textbf{1K} & \textbf{5K} & \textbf{10K} \\ \hline
\textbf{Whole Book} &  0.93    &  1.00  &  1.00 & 1.00 & 1.00  \\
\textbf{100 Sent} &   0.74   &  0.97  & 0.98  & 1.00 &  1.00 \\
\textbf{50 Sent} &   0.58   &  0.88  &   0.93 & 0.98 & 0.98 \\
\textbf{25 Sent} &  0.40  &  0.71   &  0.79 & 0.90 & 0.92 \\
\textbf{10 Sent} &  0.21  &  0.43   &  0.50 & 0.64 & 0.68 
\end{tabular}
\caption{\label{tab:test_doclen_exp50} Varying the document length in test set while it is fixed in training. Number of sentences in training documents are fixed at 50 but different lengths are considered for test documents. Numbers of sentences considered are 10, 25, 50, 100 and the entire book.  Reported results are $\bar{F}_1$ scores.}
\end{table}

\begin{figure}
    \centering
    \includegraphics[height=8cm, width=7.5cm]{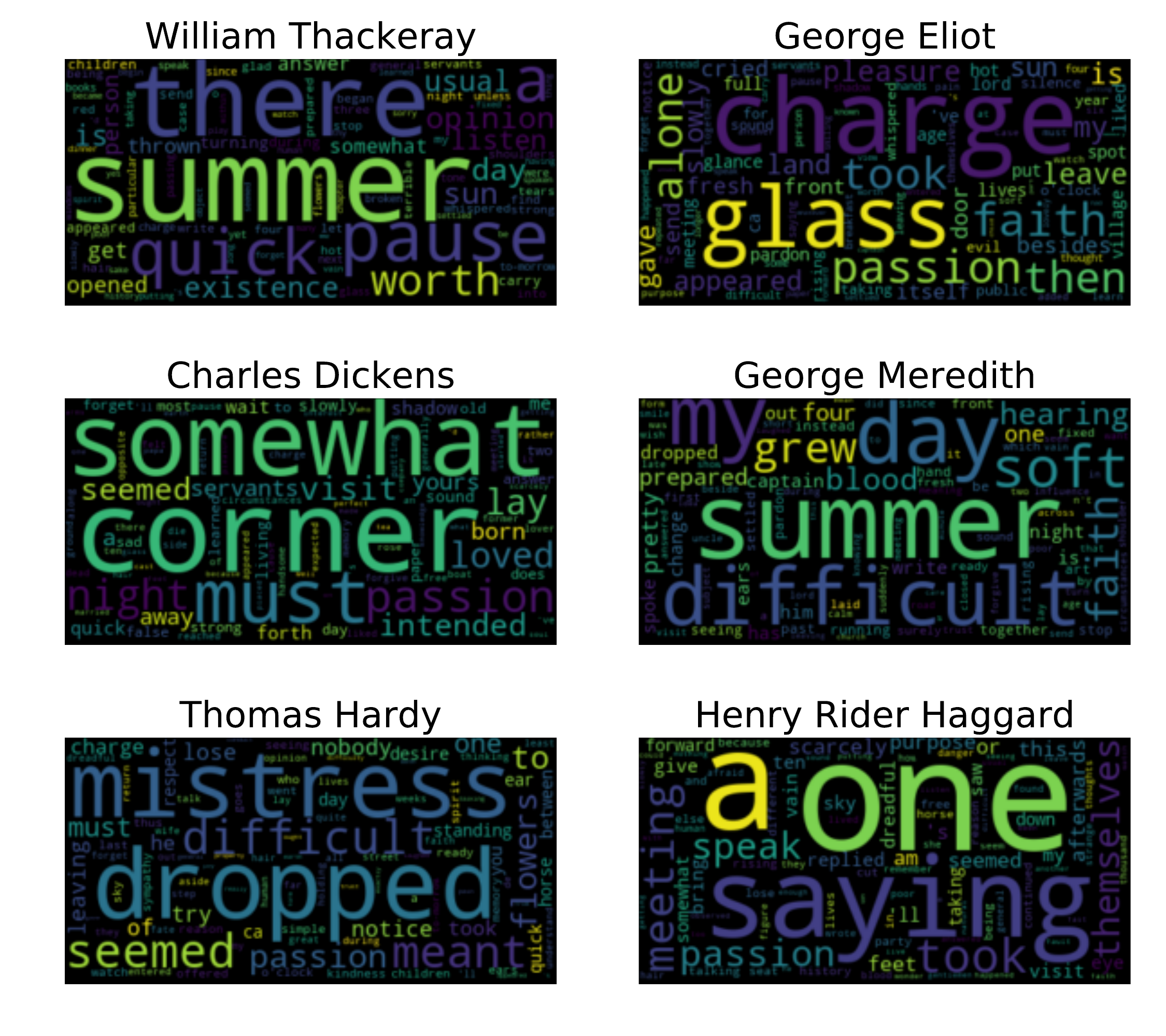}
    \caption{Word cloud plots for six authors. Words are magnified proportional to the magnitude of their corresponding SVM coefficients.}
    \label{fig:wordcloud}
\end{figure}

\textbf{Authors' top predictive words.}
This experiment is designed to highlight each author's most distinguishing words and their impact on classification. One-vs-all SVM classifiers are trained for each author with $L_1$ penalty. Documents of 50 sentences length are considered along with a vocabulary containing 1000 most frequent words. Regularization constant is tuned to maximize individual $F_1$ scores on the validation set. $L_1$ penalty yields a more sparse model by pushing coefficients of some features towards zero. From each classifier $100$ non-zero coefficients with the highest absolute magnitude are selected. We considered words associated with these coefficients as the author's most distinguishing words \footnote{Scaling  word counts eliminates the potential bias due to frequency}. Figure \ref{fig:wordcloud} displays these words for six of the authors  with font size for each word magnified proportional to the absolute magnitude of their corresponding coefficients. These six authors are Charles Dickens, George Eliot, George Meredith, H. Rider Haggard, Thomas Hardy and William Thackeray.

With one possible exception it is impractical, if not impossible, to identify authors based on these word clouds. George Eliot who was a critic of organized religion in her novels is the only exception as the word ``faith" can be spotted in the corresponding word cloud. 
%The only exception would be the top right figure where the word "faith" spotted in the word cloud can be associated with George Eliot as she was a critic of organized religion in her novels. \footnote{True matching: William Thackeray - 1, George Eliot - 2, Charles Dickens - 3, George Meredith - 4, Thomas Hardy - 5, Sir Henry Haggard - 6} 
However, the nearly-perfect attribution accuracy achieved in our results demonstrates the utility of machine learning in detecting subtle patterns not easily captured by human reasoning. In addition to presence or absence of certain words and their usage frequencies in a given text SVM takes advantage of word co-occurrence % and frequency correlations 
in the high dimensional feature space.

These word clouds provided other interesting findings as well. For example "summer" is the word with the largest absolute SVM coefficient for both William Thackeray (WT) and George Meredith (GM), yet the coefficient is positive for WT and negative for GM. This suggests that WT has its own characteristic way of using the word "summer" that would set him apart from others.\footnote{We can draw this conclusion as features from BOW representation are always non-negative. Normalization and scaling we applied did not change this fact. Therefore, we may argue that features with with the largest absolute coefficient are best individual predictors for the positive class.}. Beside word frequency distribution, correlation between these frequencies also carries predictive clues and SVM effectively exploits these signals.

The usage frequency of words with the largest absolute SVM coefficients significantly vary among authors.  For example, George Eliot's top 3 words has a total of 200 occurrence in her novels whereas H. Rider Haggard's top 3 words occur thousands of times and includes the article ``a".  

\subsection{Open-set classification}
We conducted two sets of experiments in the open set classification setting. The first one in a controlled and the other in a  real-world setting. In both settings training data contains documents from 36 known authors (64\% of author's all documents). The validation set contains documents from the first group of 5 unknown authors in addition to documents from 36 known authors (16\% of author's all documents). In the controlled setting, the test set contains documents from the second group of 5 unknown authors in addition to documents from 36 known authors (remaining 20\% of author's all documents). Document lengths are fixed at 100 sentences and most frequent $1,000$ words are used as vocabulary. In the open-set classification setting we used two different evaluation metrics as there are documents from both known and unknown authors in the test set. In addition to mean $F_1$ ($\bar{F}_1$) computed for known authors, we also report detection $F_1$ that evaluates the performance of the model in detecting documents of unknown authors. 

For the real-world experiment complete essays of various lengths from \textit{Bentley's Miscellaneous} Volume 1 and 2 (1837-1838) were considered as test documents. Numbers of sentences in these essays vary between 47 and 500.

\textbf{Results in the controlled setting.}
%Table \ref{tab:openset_controlled} reports the results of the first configuration.
$\bar{F}_1$ score  of 36 known authors in the controlled setting is $0.91$ (first row in Table \ref{tab:openset_controlled}), slightly below closed set classification score of $0.98$. This is expected as the number of potential false positives for each known author increases with the inclusion of documents from unknown authors. The first row in Table \ref{tab:openset_controlled_ood} shows the performance breakdown on OOD samples. The performance of out-of-distribution (OOD) samples detection greatly suffered from false positives (FPs), mainly because the dataset is severely unbalanced. OOD samples  form only $3.9\%$ of the test set ($365$ vs $8,949$). C. Brontë and D. M. Craik were two most affected authors with 70 and 38 FPs which were  $50\%$ and $55\%$ of total test samples from them, respectively. Small training size seems to be the underlying reason for this performance as both authors had 2 books (out of 4) for training.
%Books that have many FPs documents interestingly did not have documents falsely attributed to other known authors. They were attributed to either true author or unknown author. 
On false negative (FN) side, 3 authors were on the spotlight: C. Brontë (16), G. Meredith (12) and M. Corelli (8) where the numbers in parenthesis stand for quantity of FNs associated with that author. First row in Table \ref{tab:_misclassified_ood_together} displays author distribution over misclassified OOD samples. $14$ authors had zero false negatives and overall $33$ of them had false negatives less than or equal to five samples. Out of 365 OOD documents, our approach correctly identified 294 ($80.5\%$ ) samples belonging to an unknown author.

It is interesting to note that all $16$  documents that were incorrectly attributed to Charlotte Brontë are from her younger sister Emily Brontë's famous novel ``Wuthering Heights". Although Brontë sisters had their own narrative style and originality portrayed in their novels, they may be considered homologous when it comes to the usage of most common words. Collaborative writing and imaginary story telling during their childhood might be one explanation for this phenomenon \cite{brontes_wiki}.  
Another interesting observation is that all of the documents that were incorrectly attributed to Marie Corelli comes from Elizabeth L. Linton's book of ``Modern Woman and What is Said of Them". 

\textbf{Results on periodicals.}
Of the 27 essays collected from  Victorian periodicals, 6 are from known authors. Three of these are written under \textit{Boz} pseudonym, which is known to belong to Charles Dickens. The remaining three are written by C. Gore, G.W.M. Reynolds, and W. Thackeray, under their pseudonyms, \textit{Toby Allspy}, \textit{Max}, and \textit{Goliah Gahagan}, respectively. Remaining 21 essays are from less known writers who are not represented in our training dataset. We evaluated our results using identifications by Wellesley Index as the reference standard. Although all three articles from Charles Dickens would have been correctly classified under closed set assumption, only one was correctly attributed to him in the open set setup and the other two are misdetected as documents belonging to unknown authors. ``Adventures in Paris" under pseudonym \textit{Toby Allspy} was accurately attributed to Catherine Gore. Articles by Reynolds and Thackeray were misdetected as documents by unknown authors. 

Detection $F_1$ score of $0.84$ is still reasonable  as $18$ of the $21$ essays from unknown authors were correctly detected as documents by unknown authors.  Out of the three that were missed ``The Marine Ghost" from Edward Howard offers an interesting case study that may challenge identification by Wellesley Index. It is attributed to Frederick Marryat, who was Howard's captain while he served in the Navy. Furthermore, Marryat chose Howard as his sub-editor \cite{howard_marryat} while he was the editor of the \textit{Metropolitan Magazine}. This incorrect attribution suggests Marryat played a profound role in shaping Edward Howard's authorial voice, a claim supported by their shared naval experience and their long term professional relationship. 

The corpus has 3 essays from periodicals that still remain unattributed in Wellesley Index. In this experiment, none of three essays were attributed to any known authors. 

\begin{table}
    \centering
    \begin{tabular}{l|ccc}
        $|V|$ & \textbf{precision} & \textbf{recall} & $\boldsymbol{\bar{F}_1}$  \\ \hline
    \textbf{1K}     &  0.97 & 0.87 & 0.91 \\
    \textbf{2K} & 0.99 & 0.92 & 0.95
    \end{tabular}
    \caption{Performance on samples from known authors. Reported scores are averages from 36 authors}
    \label{tab:openset_controlled}
\end{table}

\begin{table}
    \centering
    \begin{tabular}{l|ccc}
        $|V|$ & \textbf{precision} & \textbf{recall} & $\boldsymbol{F_1}$  \\ \hline
    \textbf{1K}     &  0.34 & 0.81 & 0.48 \\
    \textbf{2K} & 0.46 & 0.85 & 0.60
    \end{tabular}
    \caption{Performance on OOD samples detection from controlled setting with 1000 and 2000 vocabulary.}
    \label{tab:openset_controlled_ood}
\end{table}

\textbf{Increasing vocabulary size }
Although the most frequent $1,000$ words proved to be sufficiently informative in the closed set classification setup, they are not as effective in the open set framework. When the number of contributors in the test set is on the order of thousands more features will inevitably be required to improve attribution accuracy. To show that this is indeed the case we repeated the previous two open set classification experiments after increasing the vocabulary size to $2,000$.

 In the controlled setting with the increased vocabulary size, $\bar{F}_1$ score on known authors increases to $0.95$ from $0.91$ while detection $F_1$ improves from $0.48$ to $0.60$  (second row in Table \ref{tab:openset_controlled}). Additional vocabulary significantly  helped to decrease total number of FPs from $569$ to $368$. Nevertheless, these features had minimal effect on documents from C. Brontë as $40\%$ of them are again classified as OOD samples.
The distribution of false negatives is also updated in the second row of  Table \ref{tab:_misclassified_ood_together}. Increasing the vocabulary size led to significant improvements in open-set classification performance yet the number of documents incorrectly attributed to Charlotte Brontë slightly increased to $18$. As earlier all of these belong to Emily Brontë's ``Wuthering Heights'' novel. This paves the way for speculating about two possible scenarios. It is possible that children with the same upbringing develop similar unconscious daily word usage, which does not change after childhood. Considering the fact that the Brontë family wrote and shared their stories with each other from a very early age \cite{charlotte_bronte}, this study suggests that their childhood editorial and reading practices shaped their subsequent work. The conflation of their respective authorial voices also suggests that the elder sister might have helped the younger in editing the book.

At the book level, using entire novel as a single long document produced perfect results. All $131$ books from known authors are attributed to their true authors and all $12$ novels from unknown authors are correctly detected as belonging to unknown authors. 
  
Noteworthy improvements are achieved on essays from periodicals as well. In addition to two articles correctly attributed to known authors in the previous experiment, article ``The Professor - A Tale" under pseudonym G. Gahagan is now correctly attributed to William Thackeray in our model. These changes also classified a previously unindexed article to Frederick Marryat. Marryat is considered an early pioneer of sea story, and the attributed article, ``A Steam Trip to Hamburg",  offers a travel narrative of a journey by sea from London to the European continent.

\begin{table}
    \centering
    \begin{tabular}{l|ccccc}
    \textbf{FN ranges} &  0 & 1 & 2-5 & 6-10 & $\geq 10$\\ \hline
   \textbf{\# authors - 1K}  & 14 & 7 & 12 & 1 & 2 \\
   \textbf{\# authors - 2K}  & 18 & 9 & 6 & 2 & 1
    \end{tabular}
    \caption{False negative distribution using most frequent $2,000$ words as a feature set. Top row represents number/range of misclassified OOD samples. The second and third rows display how many classifiers correspond to the number/range in the top row, using 1000 and 2000 most frequent words, respectively.}
    \label{tab:_misclassified_ood_together}
\end{table}

\section{Conclusion and Future Work}
In this paper, we took a pragmatic view of computational AA to highlight the critical role it could play in authorship attribution studies involving historical texts. We consider Victorian texts as a case study as many contemporary literary tropes and publishing strategies originate from this period. We demonstrated the strengths and weaknesses of existing computational AA paradigms. Specifically, we show that common English words are sufficient to a greater extent in distinguishing among writings of most renowned authors, especially when AA is performed in the closed-set setup. Experiments under closed-set assumption produced near perfect attribution accuracy in AA task involving 36 authors using only $1,000$ most frequent words. The performance suffered significantly as we switch to the more realistic open-set setup. Increasing the vocabulary size helped to some extent and provided some interesting insights that would challenge results of manual indexing. Open-set experiments also open interesting avenues for future research to investigate whether authors with the same upbringing may develop similar word usage habits as in the case of Brontë sisters. 
However, overall results from open set experiments confirm the need for a more systematic approach to open-set AA. There are several directions for future exploration following this work. 

We believe that attribution accuracy in open-set configuration could improve significantly if word counts are first mapped onto attributes capturing information about themes, genres, archetypes, settings, forms, etc. rather than being directly used in the attribution task. Bayesian priors can be extremely useful to distinguish viable human-developed word usage patterns from those adversarially generated by computers. Similarly, hierarchically clustering known authors and defining meta-authors at each level of the hierarchy can help us more accurately identify writings by unknown authors.  
The dataset that we have compiled can be enriched with additional essays from Victorian periodicals to become a challenging benchmark dataset and an invaluable resource for evaluating future computational AA algorithms.

%It can be even used to determine most similar authors to your own writing style and recommend books from those authors.

\iffalse
\subsection{Citations}
Citations within the text appear in parentheses as~\citep{Gusfield:97} or, if the author's name appears in the text itself, as \citet{Gusfield:97}.
Append lowercase letters to the year in cases of ambiguities.  
Treat double authors as in~\citep{Aho:72}, but write as in~\citep{Chandra:81} when more than two authors are involved. Collapse multiple citations as in~\citep{Gusfield:97,Aho:72}. 

\subsection{Appendices}
Appendices, if any, directly follow the text and the
references (but only in the camera-ready; see Appendix~\ref{sec:appendix}).
Letter them in sequence and provide an informative title:
\textbf{Appendix A. Title of Appendix}.

\section*{Acknowledgments}

The acknowledgments should go immediately before the references. Do not number the acknowledgments section.
Do not include this section when submitting your paper for review.
\fi

\bibliography{anthology,acl2020}

\begin{thebibliography}{38}
\expandafter\ifx\csname natexlab\endcsname\relax\def\natexlab#1{#1}\fi

\bibitem[{Alexander(1983)}]{charlotte_bronte}
Christine Alexander. 1983.
\newblock \emph{The Early Writings of Charlotte Brontë}.
\newblock Wiley, John \& Sons.

\bibitem[{Argamon and Levitan(2005)}]{argamon:05}
S.~Argamon and S.~Levitan. 2005.
\newblock Measuring the usefulness of function words for authorship
  attribution.
\newblock In \emph{Proceedings of ACH/ALLC Conference}.

\bibitem[{Argamon-Engelson et~al.(1998)Argamon-Engelson, Koppel, and
  Avneri}]{argamon_etal:98}
S.~Argamon-Engelson, M.~Koppel, and G.~Avneri. 1998.
\newblock Style-based text categorization: What newspaper am i reading?
\newblock In \emph{In Proceedings of AAAI Workshop on Learning for Text
  Categorization}, pages 1--4.

\bibitem[{Baayen et~al.(1996)Baayen, van Halteren, and
  Tweedie}]{baayen_etal:96}
R.~Baayen, H.~van Halteren, and F.~Tweedie. 1996.
\newblock Outside the cave of shadows: Using syntactic annotation to enhance
  authorship attribution.
\newblock \emph{Literary and Linguistic Computing}, 11:121–131.

\bibitem[{Bagnall(2015)}]{bagnall:15}
Douglas Bagnall. 2015.
\newblock Author identification using multi-headed recurrent neural networks.
\newblock \emph{ArXiv}, abs/1506.04891.

\bibitem[{Bozkurt et~al.(2007)Bozkurt, Baglioglu, and Uyar}]{bozkurt:07}
I.~Bozkurt, O.~Baglioglu, and E.~Uyar. 2007.
\newblock Authorship attribution performance of various features and
  classification methods.
\newblock In \emph{22nd International Symposium on Computer and Information
  Sciences}.

\bibitem[{Burrows(1992)}]{burrows:92}
J.F. Burrows. 1992.
\newblock Not unless you ask nicely: The interpretative nexus between analysis
  and information.
\newblock \emph{Literary and Linguistic Computing}, 7:91–109.

\bibitem[{Collins et~al.(2004)Collins, Kaufer, Vlachos, Butler, and
  Ishizaki}]{stylistic_incost:04}
J.~Collins, D.~Kaufer, P.~Vlachos, B.~Butler, and S.~Ishizaki. 2004.
\newblock Detecting collaborations in text: Comparing the authors’ rhetorical
  language choices in the federalist papers.
\newblock \emph{Computers and the Humanities}, 38:15--36.

\bibitem[{Curran()}]{curran_index}
Eileen Curran.
\newblock Curran index.
\newblock \url{http://curranindex.org}.

\bibitem[{Meyer~zu Eissen et~al.(2007)Meyer~zu Eissen, Stein, and
  Kulig}]{plagiarism_detection:07}
S.~Meyer~zu Eissen, B.~Stein, and M.~Kulig. 2007.
\newblock Plagiarism detection without reference collections.
\newblock \emph{Advances in Data Analysis}, pages 359--366.
\newblock Springer.

\bibitem[{Fox et~al.(2012)Fox, Ehmoda, and Charniak}]{shakespeare_dispute}
Neal Fox, Omran Ehmoda, and Eugene Charniak. 2012.
\newblock Statistical stylometrics and the marlowe- shakespeare authorship
  debate.
\newblock Providence, RI: Brown University M.A Thesis.

\bibitem[{Gamon(2004)}]{gamon:04}
M.~Gamon. 2004.
\newblock Linguistic correlates of style: Authorship classification with deep
  linguistic analysis features.
\newblock In \emph{In Proceedings of the 20th International Conference on
  Computational Linguistics}, pages 611--617.

\bibitem[{GDELT()}]{gdelt}
Project GDELT.
\newblock \url{https://www.gdeltproject.org}.

\bibitem[{Goodwin(1885)}]{howard_marryat}
Gordon Goodwin. 1885.
\newblock Howard, edward.
\newblock \emph{Dictionary of National Biography}, 28.

\bibitem[{Gungor(2018)}]{mecit_thesis:18}
A.~Gungor. 2018.
\newblock Benchmarking authorship attribution techniques using over a thousand
  books by fifty victorian era novelists.
\newblock \emph{Master Thesis}.

\bibitem[{Gutenberg()}]{gutenberg}
Project Gutenberg.
\newblock \url{https://www.gutenberg.org}.

\bibitem[{Haberfeld and Hassell(2009)}]{unibomber}
M.~Haberfeld and A.~V. Hassell. 2009.
\newblock A new understanding of terrorism: Case studies, trajectories and
  lessons learned.
\newblock \emph{Springer}.

\bibitem[{Hitschler et~al.(2017)Hitschler, van~den Berg, and
  Rehbein}]{julian_etal:17}
Julian Hitschler, Esther van~den Berg, and Ines Rehbein. 2017.
\newblock Authorship attribution with convolutional neural networks and
  pos-eliding.
\newblock In \emph{In Proceedings of the Workshop on Stylistic Variation}, page
  53–58.

\bibitem[{Holmes(1998)}]{holmes:98}
D.I. Holmes. 1998.
\newblock The evolution of stylometry in humanities scholarship.
\newblock \emph{Literary and Linguistic Computing}, 13:111--117.

\bibitem[{Hu(2013)}]{new_testament:13}
W.~Hu. 2013.
\newblock Study of pauline epistles in the new testament using machine
  learning.
\newblock \emph{Sociology Mind}, 3:193--203.

\bibitem[{Kim et~al.(2010)Kim, Kim, Weninger, and Han}]{kim_etal:10}
S.~Kim, H.~Kim, T.~Weninger, and J.~Han. 2010.
\newblock Authorship classification: Syntactic tree mining approach.
\newblock In \emph{Proceedings of the ACM SIGKDD Workshop on Useful Patterns}.

\bibitem[{Koppel et~al.(2002)Koppel, Argamon, and
  Shimoni}]{author_profiling:02}
M.~Koppel, S.~Argamon, and A.R. Shimoni. 2002.
\newblock Automatically categorizing written texts by author gender.
\newblock \emph{Literary and Linguistic Computing}, 17:401--412.

\bibitem[{Koppel and Schler(2004)}]{koppel_schler:04}
M.~Koppel and J.~Schler. 2004.
\newblock Authorship verification as a one-class classification problem.
\newblock In \emph{Proceedings of the 21st International Conference on Machine
  Learning}.

\bibitem[{Koppel and Winter(2014)}]{koppel:14}
Moshe Koppel and Yaron Winter. 2014.
\newblock Determining if two documents are written by the same author.
\newblock \emph{Journal of the Association for Information Science and
  Technology}, 65:178–187.

\bibitem[{McCarthy et~al.(2006)McCarthy, Lewis, Dufty, and
  McNamara}]{McCarthy_etal:06}
P.M. McCarthy, G.A. Lewis, D.F. Dufty, and D.S. McNamara. 2006.
\newblock Analyzing writing styles with coh-metrix.
\newblock In \emph{In Proceedings of the Florida Artificial Intelligence
  Research Society International Conference}, pages 764--769.

\bibitem[{Mendenhall(1887)}]{mendenhall:87}
T.~C Mendenhall. 1887.
\newblock The characteristic curves of composition.
\newblock \emph{Science}, 9:237–249.

\bibitem[{Morton and Michaelson(1990)}]{morton_et_al:90}
A.Q. Morton and S.~Michaelson. 1990.
\newblock The qsum plot.
\newblock Technical Report CSR-3-90, University of Edinburgh.

\bibitem[{Mosteller and Wallace(1964)}]{mostellar:64}
F.~Mosteller and D.L. Wallace. 1964.
\newblock Inference and disputed authorship: The federalist.
\newblock Addison-Wesley.

\bibitem[{Olsson(2008)}]{ollson:08}
J.~Olsson. 2008.
\newblock Forensic linguistics: An introduction to language, crime and the law.
\newblock 2nd ed.

\bibitem[{Peng et~al.(2004)Peng, Shuurmans, and Wang}]{peng_etal:04}
F.~Peng, D.~Shuurmans, and S.~Wang. 2004.
\newblock Augmenting naive bayes classifiers with statistical language models.
\newblock \emph{Information Retrieval Journal}, 7:317--345.

\bibitem[{Rhodes(2015)}]{rhodes:15}
D.~Rhodes. 2015.
\newblock \href {http://cs224d.stanford.edu/reports/RhodesDylan.pdf} {Author
  attribution with cnns}.
\newblock Technical Report.

\bibitem[{Rudman(1998)}]{rudman:98}
J.~Rudman. 1998.
\newblock The state of authorship attribution studies: Some problems and
  solutions.
\newblock \emph{Computers and the Humanities}, 31:351--365.

\bibitem[{Scheirer et~al.(2013)Scheirer, Rocha, Sapkota, and
  Boult}]{openset_recognition:12}
W.~J. Scheirer, A.~R. Rocha, A.~Sapkota, and T.~E. Boult. 2013.
\newblock Toward open set recognition.
\newblock \emph{TPAMI}, 35.

\bibitem[{Stamatatos(2009)}]{Stamatatos:09}
Efstathios Stamatatos. 2009.
\newblock A survey of modern authorship attribution methods.
\newblock \emph{Journal of the American Society for Information Science and
  Technology}, 60:538--556.

\bibitem[{Thompson and Rasp(2009)}]{dark_tower:13}
J.~R. Thompson and J.~Rasp. 2009.
\newblock Did c.s. lewis write the dark tower?: An examination of the
  small-sample properties of the thisted-efron tests of authorship.
\newblock \emph{Austrian Journal Statistics}, 38:71--82.

\bibitem[{de~Vel et~al.(2001)de~Vel, Anderson, Corney, and
  Mohay}]{deVel_etal:01}
O.~de~Vel, A.~Anderson, M.~Corney, and G.~Mohay. 2001.
\newblock Mining e-mail content for author identification forensics.
\newblock \emph{SIGMOD Record}, 30:55--64.

\bibitem[{Wellesley()}]{wellesley_index}
Index Wellesley.
\newblock \url{http://wellesley.chadwyck.co.uk}.

\bibitem[{Wikipedia(2019)}]{brontes_wiki}
Wikipedia. 2019.
\newblock Brontë family.
\newblock Https://en.wikipedia.org/wiki/Brontë\_family.

\end{thebibliography}
\bibliographystyle{acl_natbib}

\iffalse
\appendix

\section{Appendices}
\label{sec:appendix}
Appendices  

\section{Supplemental Material}
\label{sec:supplemental}
Supp. Material
\fi

\end{document}